\documentclass[10pt,twocolumn,letterpaper]{article}

\usepackage{iccv}
\usepackage{times}
\usepackage{epsfig}
\usepackage{graphicx}
\usepackage{amsmath}
\usepackage{amssymb}
\usepackage{subfig}
\usepackage{multirow}
\usepackage[symbol]{footmisc}
\usepackage{url}
\def\etc{\emph{etc}}
\def\etal{\emph{et~al}.}

\usepackage[breaklinks=true,bookmarks=false]{hyperref}

\iccvfinalcopy 

\def\iccvPaperID{2402} 

\ificcvfinal\fi
\begin{document}

\title{Chinese Street View Text: Large-scale Chinese Text Reading with Partially Supervised Learning}

\author{Yipeng Sun$^1$ \quad Jiaming Liu$^1$ \quad  Wei Liu$^2$\thanks{This work is done when Wei Liu is an intern at Baidu Inc.} \quad Junyu Han$^1$ \quad Errui Ding$^1$ \quad Jingtuo Liu$^1$\\
Department of Computer Vision Technology (VIS), Baidu Inc.$^1$ \\
\quad Department of Computer Science, The University of Hong Kong$^2$\\
{\tt\small \{sunyipeng,liujiaming03,hanjunyu,dingerrui,liujingtuo\}@baidu.com}\\
{\tt\small wliu@cs.hku.hk}
}

%

\maketitle
\ificcvfinal\thispagestyle{empty}\fi

\begin{abstract}
Most existing text reading benchmarks make it difficult to evaluate the performance of more advanced deep learning models in large vocabularies due to the limited amount of training data. To address this issue, we introduce a new large-scale text reading benchmark dataset named \textbf{C}hinese \textbf{S}treet \textbf{V}iew \textbf{T}ext~(C-SVT) with $430,000$ street view images
, which is at least $\bf{14}$ times as large as the existing Chinese text reading benchmarks.
To recognize Chinese text in the wild while keeping large-scale datasets labeling cost-effective, we propose to annotate one part of the C-SVT dataset (30,000 images) in locations and text labels as full annotations and add $400,000$ more images, where only the corresponding text-of-interest in the regions is given as weak annotations. To exploit the rich information from the weakly annotated data, we design a text reading network in a partially supervised learning framework, which enables to localize and recognize text, learn from fully and weakly annotated data simultaneously. To localize the best matched text proposals from weakly labeled images, we propose an online proposal matching module incorporated in the whole model, spotting the keyword regions by sharing parameters for end-to-end training. Compared with fully supervised training algorithms, this model can improve the end-to-end recognition performance remarkably by $\bf{4.03\%}$ in F-score at the same labeling cost. The proposed model can also achieve state-of-the-art results on the ICDAR 2017-RCTW dataset, which demonstrates the effectiveness of the proposed partially supervised learning framework.
\end{abstract}
\vspace{-0.8em}
\section{Introduction}
Reading text from images has received much attention in recent years due to its numerous applications, e.g., document analysis, image-based translation, product image retrieval, visual geo-location and license plate recognition, etc. Benefitting from the advances in deep learning algorithms~\cite{he2016deep, ren2015faster, liu2016ssd, graves2013speech, bahdanau2014neural}, the performance of text detection and recognition on standard benchmarks has increased dramatically over the past three years~\cite{icdar15-web} .
\begin{figure}[!ht]
\centering
\includegraphics[width=0.495\linewidth]{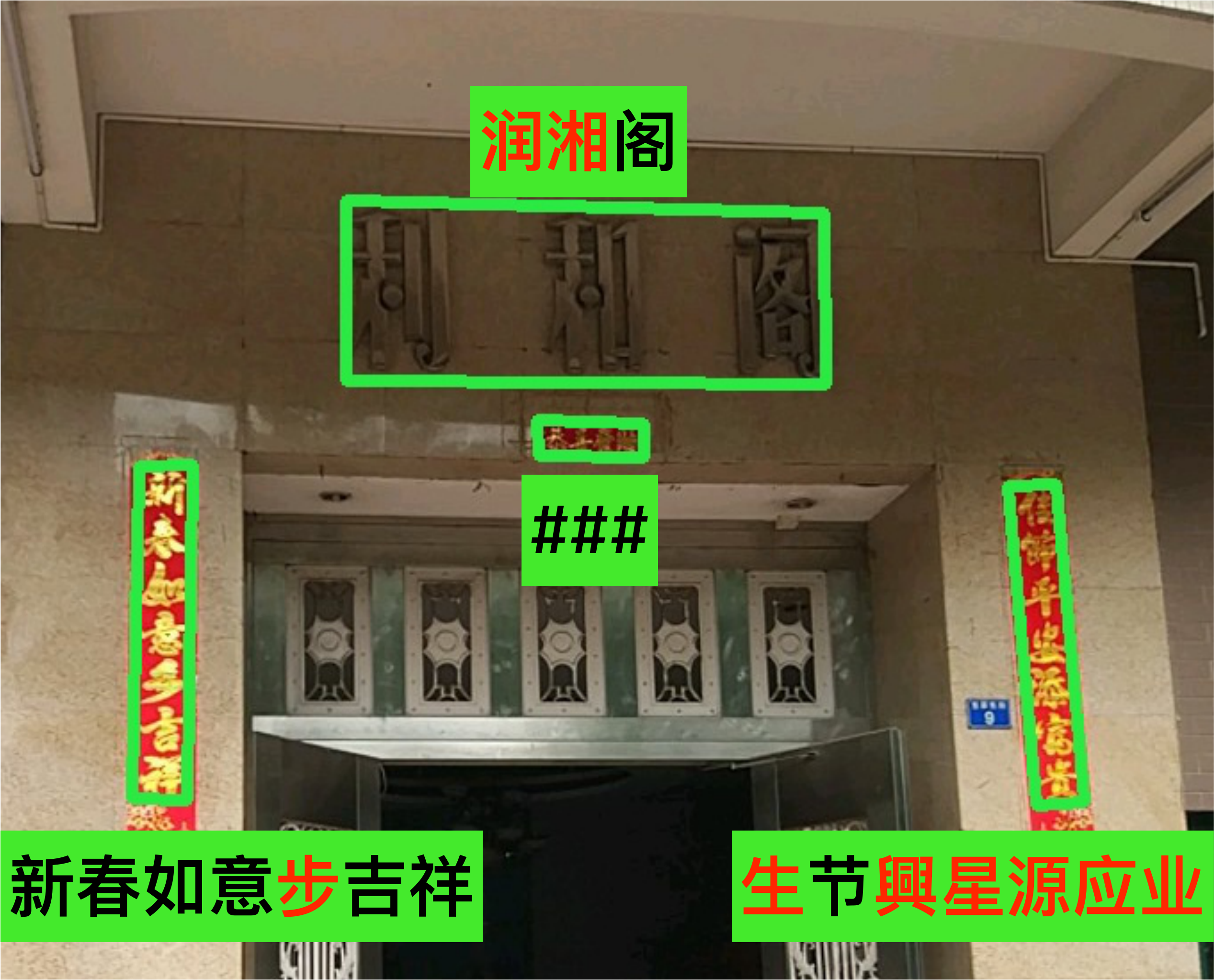}
\includegraphics[width=0.495\linewidth]{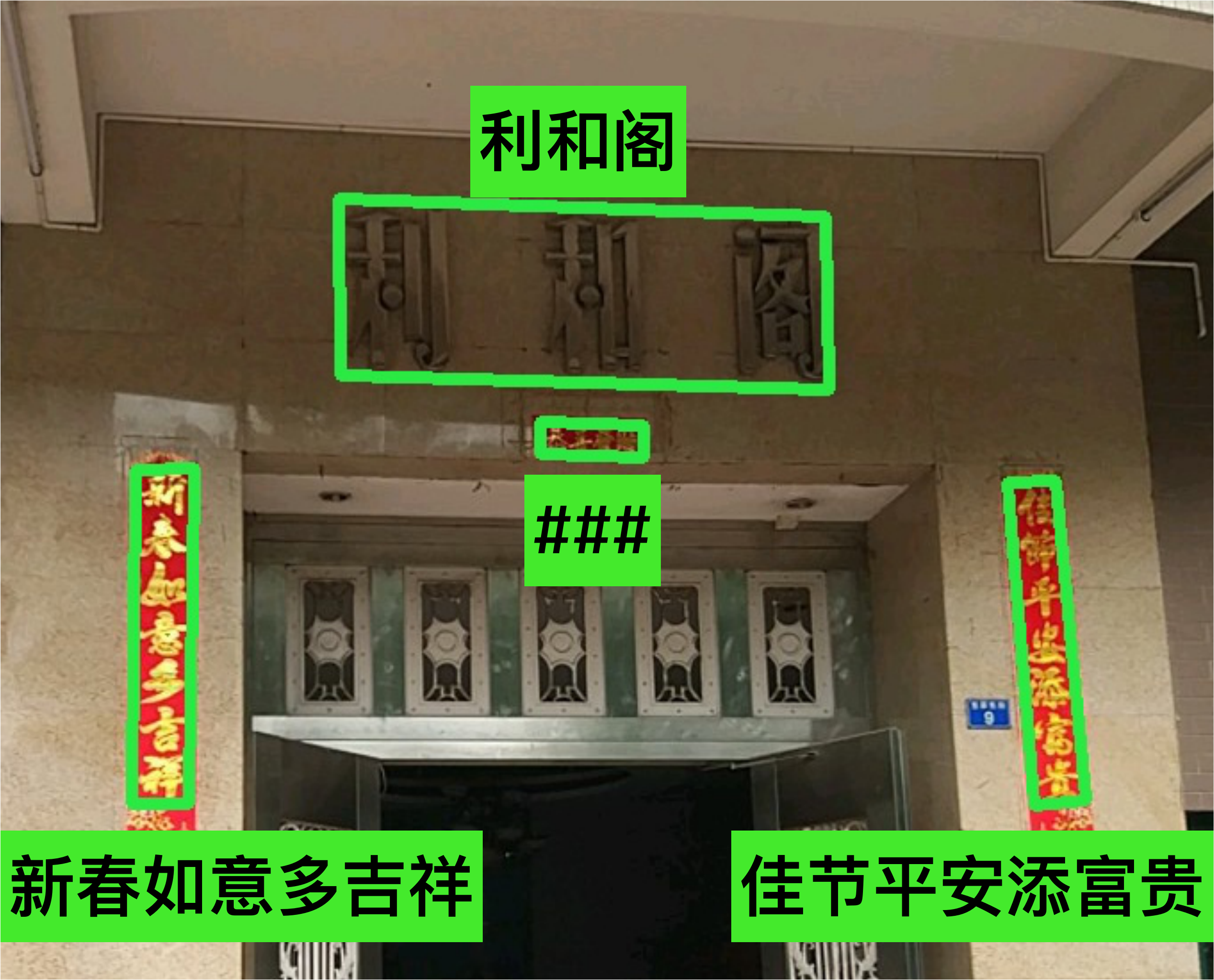}\vspace{-0.5em}
\caption{End-to-end Chinese text reading: fully supervised~(left) vs.~partially supervised~(right). Note that the characters recognized incorrectly are marked in red.}\label{figure:intro}\vspace{-0.75em}
\end{figure} 
Thanks to the carefully designed deep learning based text reading models, remarkable success has been achieved in terms of detecting~\cite{zhou2017east, shi2017detecting, 2017wordsup} and recognizing text~\cite{shi2017end, shi2016robust, lee2016recursive, cheng2017focusing} on ICDAR benchmarks \cite{karatzas13icdar, karatzas15icdar}, which mainly focus on English text in the wild. 

However, previous approaches have rarely paid attention to reading Chinese text in the wild. There is a considerable drop in performance when applying the state-of-the-art text detection and recognition algorithms to Chinese text reading, which is more challenging to solve. Since the category number of Chinese characters in 
real-world images is much larger than those of Latin languages, the number of training samples of most current datasets is still limited per category and the distribution of characters is relatively unbalanced. Therefore, reading Chinese text in the wild requires more well annotated training samples, however, it is difficult for the existing benchmarks~\cite{shi2017icdar2017}\cite{yuan2018chinese} to satisfy the requirements mainly due to the high cost of data collections and location annotations of text regions. 

To address this issue, we establish a new large-scale Chinese text reading benchmark named {\bf C}hinese {\bf S}treet {\bf V}iew {\bf T}ext ({\bf C-SVT}) with more than $430,000$ street view images in total. This dataset contains $30,000$ fully annotated images with locations and text labels for the regions. Since annotating images in precise locations of text regions is extremely time-consuming, expensive and inefficient for practical applications, we add $400,000$ more images in which only the annotations of the text-of-interest are given. These great numbers of images are much cheaper to collect and annotate, referred to as weak annotations.  
In this work, we intend to exploit such great numbers of weakly labeled images by a novel solution in a single model. Specifically, we propose an end-to-end trainable Chinese text reading model in a partially supervised learning framework. To localize the best matched text proposals from weakly labeled images, we develop an Online Proposal Matching module (OPM) integrated in the framework. This model can primarily provide end-to-end localization and recognition results using full annotations and further substantially boost the performance via partially supervised joint training, making use of both full and weak annotations in a unified framework.

The contributions of the paper are three-fold. \textbf{1)} We establish a new large-scale Chinese text reading benchmark named C-SVT, providing full and weak text annotations, which is $14$ times as large as the existing Chinese text reading datasets. \textbf{2)} To exploit large-scale weak annotations, we propose an partially supervised end-to-end trainable text reading model, which enables to learn from both full and weak annotations, simultaneously localize and recognize Chinese text in wild. \textbf{3)} The performance of the proposed partially supervised model can remarkably surpass that of the fully supervised learning model and achieve the state-of-the-art results on ICDAR 2017-RCTW as well.


To the best of our knowledge, the proposed C-SVT dataset is the largest Chinese text reading benchmark to date. It is also the first time that partially supervised learning is proposed to tackle the end-to-end text reading task.

\vspace{-0.5em}
\section{Related work}
\subsection{Text Reading Benchmarks}
Recently, many datasets have been collected for reading text in natural images, which contributes greatly to the advances in latest text reading methods. For English text, ICDAR 2013 \cite{karatzas13icdar} and ICDAR 2015 \cite{karatzas15icdar} which mainly contain horizontal and multi-orientated text, are first utilized to evaluate the performance of text reading models. In order to handle the text in more complicated situations, the curved text is introduced in Total-Text~\cite{chng17tt} and SCUT-CTW1500 \cite{yuliang2017detecting} datasets. For Chinese text, Liu \etal~\cite{liu2011casia} first introduced a dataset for the online and offline handwritten recognition. For Chinese text in the wild, MSRA-TD500~\cite{yao2012detecting}, RCTW-17~\cite{shi2017icdar2017} and CTW~\cite{yuan2018chinese} have been released to evaluate the performance of Chinese text reading models. Unlike all the previous datasets which only provide fully annotated images, the proposed C-SVT dataset also introduces a large amount of weakly annotated images with only the text labels in regions-of-interest, which are much easier to collect and have the potential to further improve the performance of text reading models. C-SVT is at least $14$ times as large as the previous Chinese benchmarks~\cite{shi2017icdar2017,yuan2018chinese}, making it the largest dataset for reading Chinese text in the wild.

\subsection{End-to-End Text Reading}
End-to-end text reading has received much attention as numerous practical applications can benefit from the rich semantic information embedded in natural images. Most of the traditional methods \cite{wang2012end,gupta2016synthetic,jaderberg2016reading,liao2018textboxes++} split this task into two separate parts. They first employ a detector to localize text regions in the images and then generate characters in the detected regions by a text recognizer. To jointly optimize these two parts by sharing features, recent methods \cite{li2017towards,2017deep_textspotter,bartz2017stn,bartz2018see,liu2018fots,he2018end,lyu2018mask,sun2018textnet} employ an end-to-end trainable framework to localize and recognize text regions simultaneously. For the detection branch, \cite{buvsta2017deep,li2017towards,lyu2018mask} utilized a region proposal network to generate text proposals and \cite{he2018end,liu2018fots,sun2018textnet} adopted a fully convolutional network to directly predict locations \cite{zhou2017east}. For the recognition branch, the CTC (Connectionist Temporal Classification)~\cite{2017deep_textspotter,liu2018fots} and attention-based LSTM decoder \cite{li2017towards,he2018end,sun2018textnet} were used to recognize individual characters as sequence-to-sequence problems~\cite{graves2013speech, bahdanau2014neural}. Different from all the previous end-to-end text reading models which are trained in a fully supervised manner, our model is trained with a partially supervised learning framework. By incorporating the large-scale weakly annotated data into the training process, we can further improve the end-to-end performance by a large margin.

\subsection{Weakly and Partially Supervised Learning}
Weakly supervised learning has been used in many areas of computer vision, e.g., image recognition \cite{mahajan2018exploring}, segmentation \cite{Hu_2018_CVPR,shen2018bootstrapping} and scene text detection \cite{tian2017wetext,2017wordsup}, \etc. Weakly supervised scene text detection methods~\cite{tian2017wetext,2017wordsup} trained a supervised character-based detector using character-level bounding box and refined this model with word-level bounding box to improve the accuracy. In the end-to-end text reading task, we exploit large-scale weakly labeled data together with fully labeled data to train our text reading model, which is formulated as a partially supervised learning framework \cite{Hu_2018_CVPR} to simultaneously localize and recognize text regions. To the best of our knowledge, it is the first time that a partially supervised learning paradigm is introduced into the end-to-end text reading task. 

\section{Chinese Street View Text}
In this section, we introduce the Chinese Street View Text benchmark and the characteristics of the dataset, including full and weak annotations. 

\subsection{Definitions}
The Chinese Street View Text benchmark aims to evaluate more advanced deep learning models with the help of more than $430,000$ image samples, which is over $14$ times as large as the existing Chinese text benchmarks, as listed in Tab. \ref{table:datasets_compare} for comparisons.  All images in the C-SVT dataset are collected from the real streets in China. As annotating ground truths in precise text locations is expensive and time-consuming, it is inefficient to collect and label such large quantities in precise locations with text labels. 
\begin{table*}[!ht]
\centering
\renewcommand{\arraystretch}{1.1}
\caption{Comparisons among existing text reading benchmarks.}\label{table:datasets_compare}\vspace{-0.8em}
\scriptsize
\begin{tabular}{|c|c|c|c|c|c|c|c|c}
\hline
Dataset        & Year  & Scenario      & Major language & Number of Images  & Label \\ 

\hline\hline
MSRA-TD500~\cite{yao2012detecting}  & 2012  & Street view   & Chinese     & 500     & Quadrangles \\ 
\hline

ICDAR 2013~\cite{karatzas13icdar}  & 2013  & Focus scene   & English       & 500+     & Quadrangles + Words \\ 

\hline
ICDAR 2015~\cite{karatzas15icdar}  & 2015  & Incidental scene  & English       & 1.5K     & Quadrangles + Words \\

\hline
DOST~\cite{iwamura2016downtown}          & 2016  & Video sequence & Japanese  & 30K+ frames   & Quadrangles + Words  \\

\hline
COCO-Text~\cite{veit2016coco}  & 2016  & Incidental scene & English     & 60K+   & Rectangles + Words  \\

\hline
Total-Text~\cite{chng17tt}  & 2017  & Incidental scene  & English & 1.5K+  & Polygons + Words \\

\hline
Uber-Text~\cite{UberText}  & 2017  & Street view & English  & 110K  & Polygons + Words  \\

\hline
ICDAR-17 MLT~\cite{icdar2017-mlt}  & 2017  & Incidental scene  & Multilingual  & 10K+   & Quadrangles + Words/Text lines \\

\hline
ICDAR 2017-RCTW~\cite{shi2017icdar2017} & 2017  & Street view + web images  & Chinese  & 12K+   & Quadrangles + Words/Text lines  \\

\hline
CTW~\cite{yuan2018chinese} & 2018  & Street view  & Chinese  & 30K   & Rectangles + Characters  \\ 

\hline
ICPR 2018-MTWI~\cite{icpr-2018-mtwi} & 2018  & Web images  & Chinese & 20K  & Quadrangles + Words/Text lines \\ 
\hline\hline
C-SVT & 2019  & Street view  & Chinese & 30K + 400K  & Quadrangles/Polygons + Words/Text lines \\ 
\hline
\end{tabular}\vspace{-1.4em}
\end{table*}

To address this issue, we propose to annotate images in two separate parts. One part of samples is labeled in locations of text regions with the corresponding text labels, referred to as full annotations. The other part is annotated in rough mask regions and only the labels of text-of-interest are given as weak annotations. Note that text-of-interest is referred to as the meaningful keyword information labeled by users. Fully supervised learning is to train models with fully annotated data, and weakly supervised learning is applied on the weakly annotated data instead. Training on the mixture of fully and weakly annotated data is referred to as partially supervised learning~\cite{mahajan2018exploring}.

\begin{figure}[t]
   \begin{center}
      \includegraphics[width=0.523\linewidth]{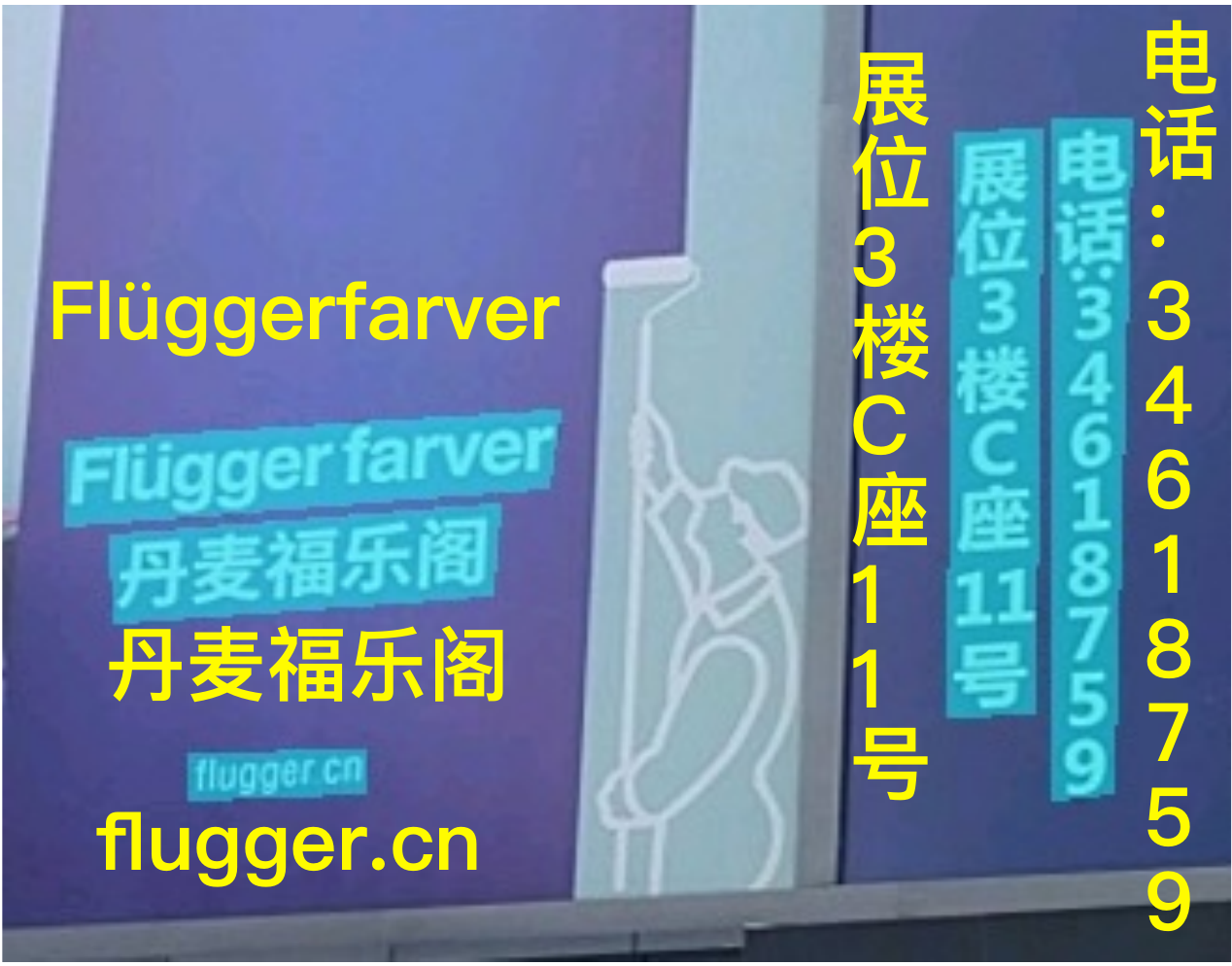}
      \includegraphics[width=0.41\linewidth]{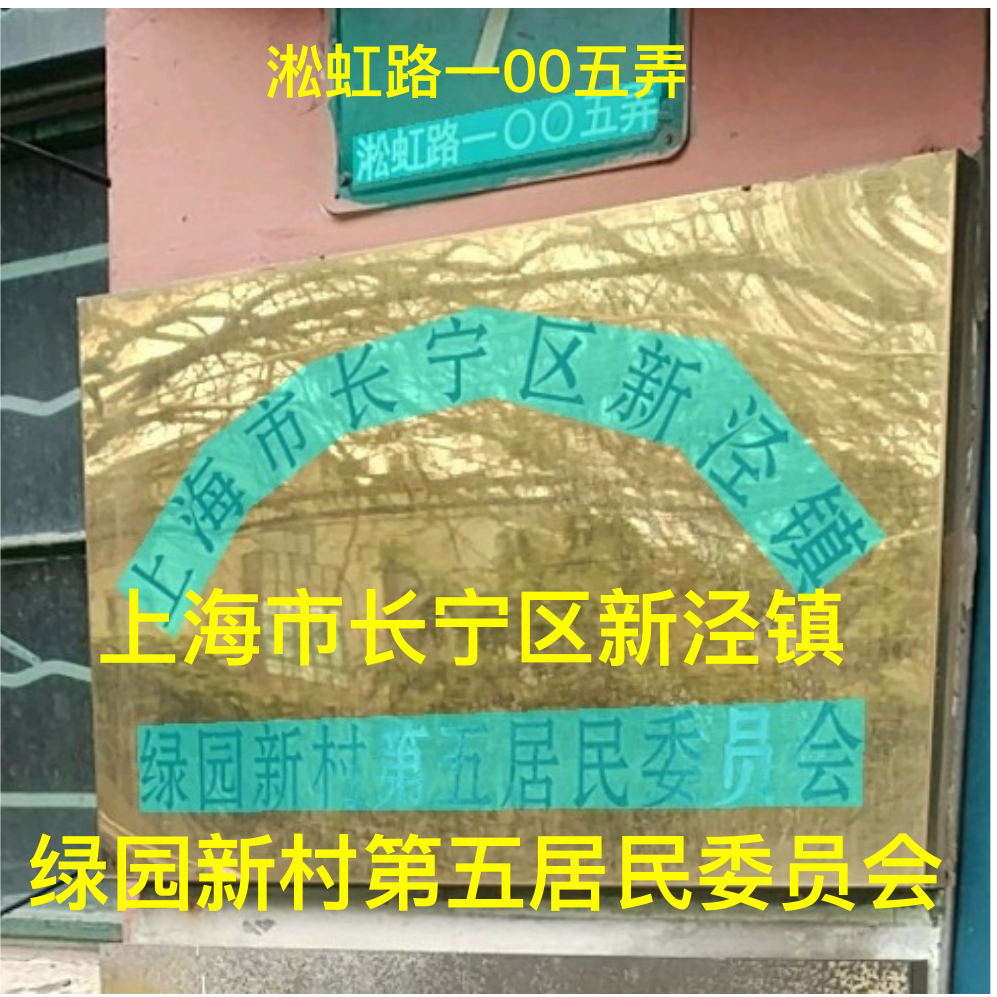}
   \end{center}\vspace{-1.5em}
\caption{\small Examples in full annotations. Characters in yellow include Chinese, numbers and alphabets. Text regions in green include horizontal, vertical and oriented cases in quadrangles. Curved text lines are marked in multiple points.}\label{figure:exmaples_fad}\vspace{-1.2em}
\end{figure}

\vspace{-0.8em}
\paragraph{Full Annotations:} There are $29,966$ fully annotated images in which the locations and text labels in the corresponding regions are given. The annotated text instances include horizontal, vertical and perspective as well as curved cases, which are represented in quadrangles or polygons as shown in Fig.~\ref{figure:exmaples_fad}. We split the fully annotated dataset into a training set, a validating set and a testing set, setting the size ratio of the three sets approximately to $4 : 1 : 1$. The whole splitting procedure is applied to keep horizontal and vertical text regions as well as different Chinese character categories almost equally distributed in the three sets. The fully annotated dataset contains $29,966$ images with $243,537$ text lines and $1,509,256$ characters in total. The detailed statistics of training, validation and test sets after splitting are shown in Tab. \ref{table:full_statistics}. This includes the numbers of horizontal, vertical and curved text lines, numbers of characters and Chinese for the three sets, respectively.
\begin{table}[t]
\tabcolsep=0.3em
\centering
\renewcommand{\arraystretch}{1.1}
\caption{\small Statistics of the training, validation and test sets in full annotations. Note that the ratio between horizontal and vertical text lines is kept to $6:1$ in all dataset splits.}\label{table:full_statistics}\vspace{-0.5em}
\scriptsize
\begin{tabular}{|c|c|c|c|c|c|c|c|c}
\hline
\multirow{2}*{Split}         & \multirow{2}*{Image number} & \multicolumn{3}{c|}{Horizontal vs. Vertical vs. Curved}  & \multicolumn{2}{c|}{Numbers}  \\ 
\cline{3-7}
                             &                              &  Horizontal & Vertical &  Curved &           Characters          & Chinese          \\
\hline
Train                        & 20,157                        & 139,613     & 23,473   & 2,471    & 1,031,904               & 620,368     \\
\hline
Valid                        & 4,968                         & 33,019       & 5,709   & 600       & 240,261                & 144,088     \\
\hline
Test                         & 4,841                         & 32,470       & 5,575   & 607       & 237,091                & 143,849     \\
\hline
Total                        & 29,966                        & 205,102      & 34,757   & 3,678     & 1,509,256               & 908,305     \\
\hline
\end{tabular}\vspace{-1.2em}
\end{table}

\begin{figure}[t]
     \begin{center}
      \includegraphics[width=\linewidth]{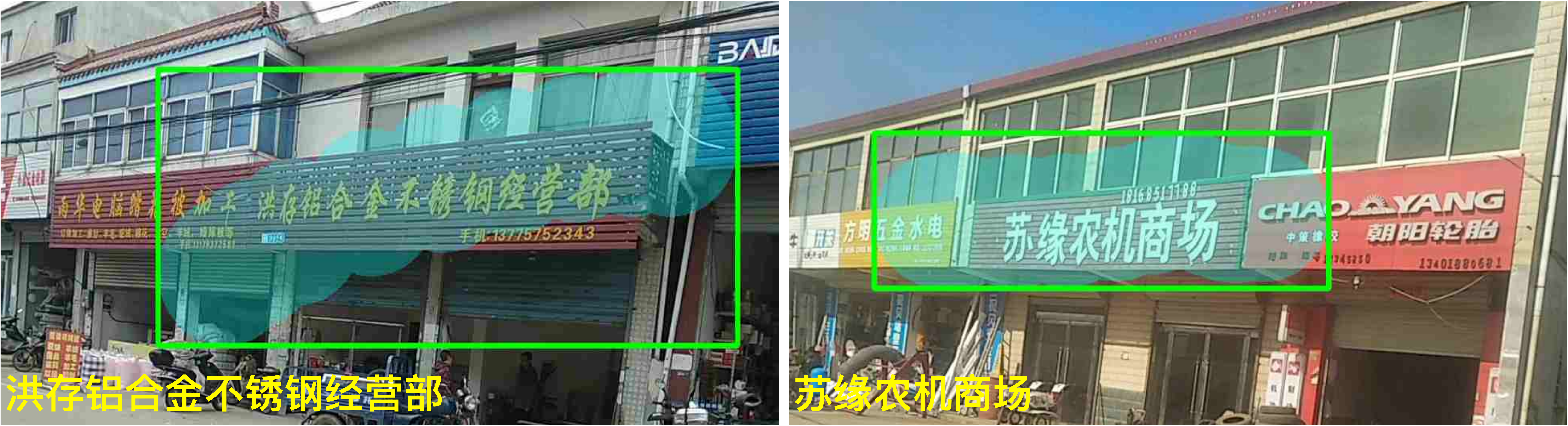}
     \end{center}\vspace{-1.5em}
  \caption{\small Examples in weak annotations. Note that the locations of text-of-interest are labeled by hand-crafted masks as visualized in green, and the corresponding characters of keywords are labeled in yellow at the bottom left corner.}\label{figure:exmaples_pad}\vspace{-1.2em}
\end{figure}

\vspace{-0.8em}
\paragraph{Weak Annotations:} It is difficult to use previous text reading benchmarks to evaluate more powerful deep learning models due to the lack of training samples, especially for Chinese text reading in large vocabularies. To overcome this limitation while keeping labeling cost-effective, we add $400,000$ more images in which only the annotations of text-of-interest are given with rough locations and text labels, as shown in Fig.~\ref{figure:exmaples_pad}. The weakly annotated text-of-interest include $5$ million Chinese characters in total. These weak annotations can be further utilized to boost the end-to-end recognition performance, while the collection and labeling jobs for such great numbers of image samples are much easier and more efficient. 



  
\subsection{Data Collection and Labeling}
The whole procedure to construct the dataset mainly includes data collection, pre-processing and labeling. First, these images are captured by mobile phones by crowd-sourcing in the streets of China across different cities. Then we utilize the face and license plate detection algorithms to address privacy issues by blurring the detected regions. The labeling process is conducted on a developed crowd-sourcing platform following the annotation definitions. As a labor intensive job, the labeling process takes $55$ man in three weeks, i.e., $6,600$ man-hours, to complete $30$K full annotations. This in contrast to the time and efforts required for $400$K weak annotations, which only required $960$ man-hours ($24$ individuals, one week). The labeling cost of $30$K full annotations is approximately $6.88$ times as much as that of $400$K weak annotations. 
Finally, we verify the labeling quality of the datasets to guarantee that the labeling error rates are controlled to no more than $2\%$ . To check the labeling quality, several annotators randomly select $30$\% chunks of labeled data to check whether the accuracy is above $98$\% otherwise the data will be annotated again.

\subsection{Evaluation Metrics}
We evaluate the performance of models on the C-SVT dataset in text detection and end-to-end recognition tasks. Following the evaluation rules of ICDAR 2015~\cite{karatzas15icdar} and ICDAR 2017-RCTW~\cite{shi2017icdar2017}, the text detection task of C-SVT is evaluated in terms of Precision, Recall and F-score when the IoU~(intersection-over-union) is above $0.5$. To compare results more comprehensively, the end-to-end performance of C-SVT is evaluated from several aspects including AED~(Average Edit Distance)
~\cite{shi2017icdar2017}, Precision, Recall and F-score. Under the exactly matched criteria in F-score, a true positive text line means that the Levenshtein distance between the predicted result and ground-truth equals to zero when IoU is above $0.5$.

\begin{figure*}[t]
   \begin{center}
      \includegraphics[width=0.64\linewidth]{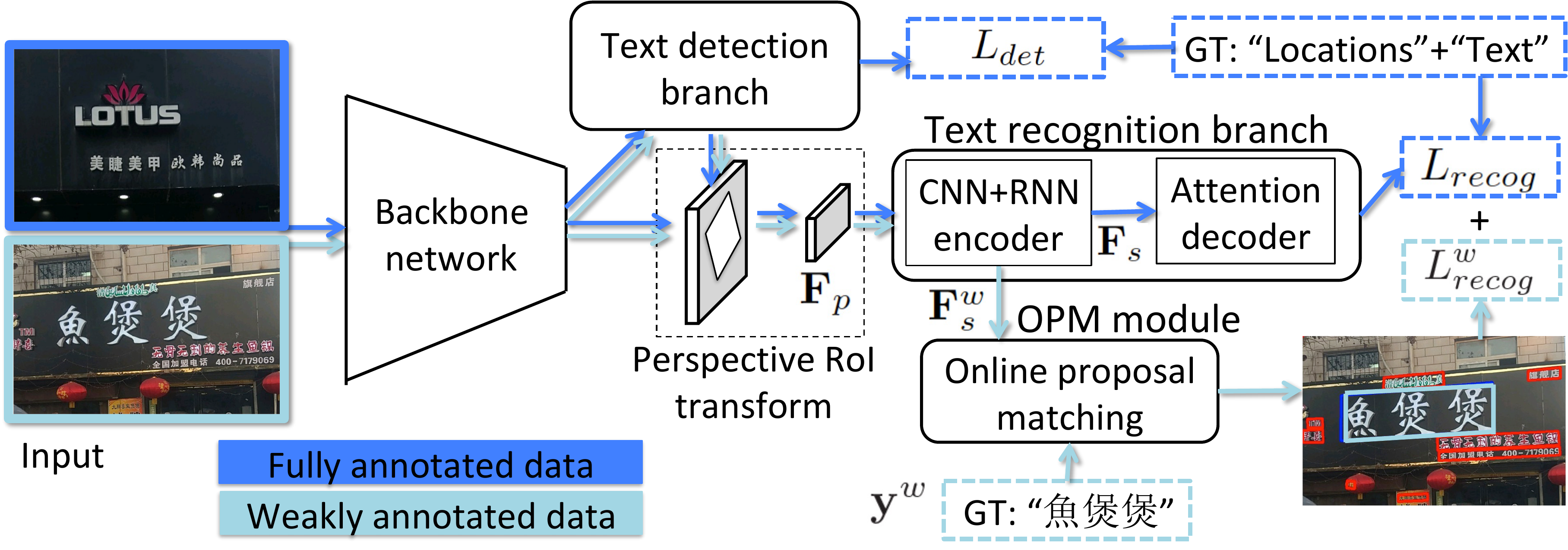}~~~~~~
     \includegraphics[width=0.32\linewidth]{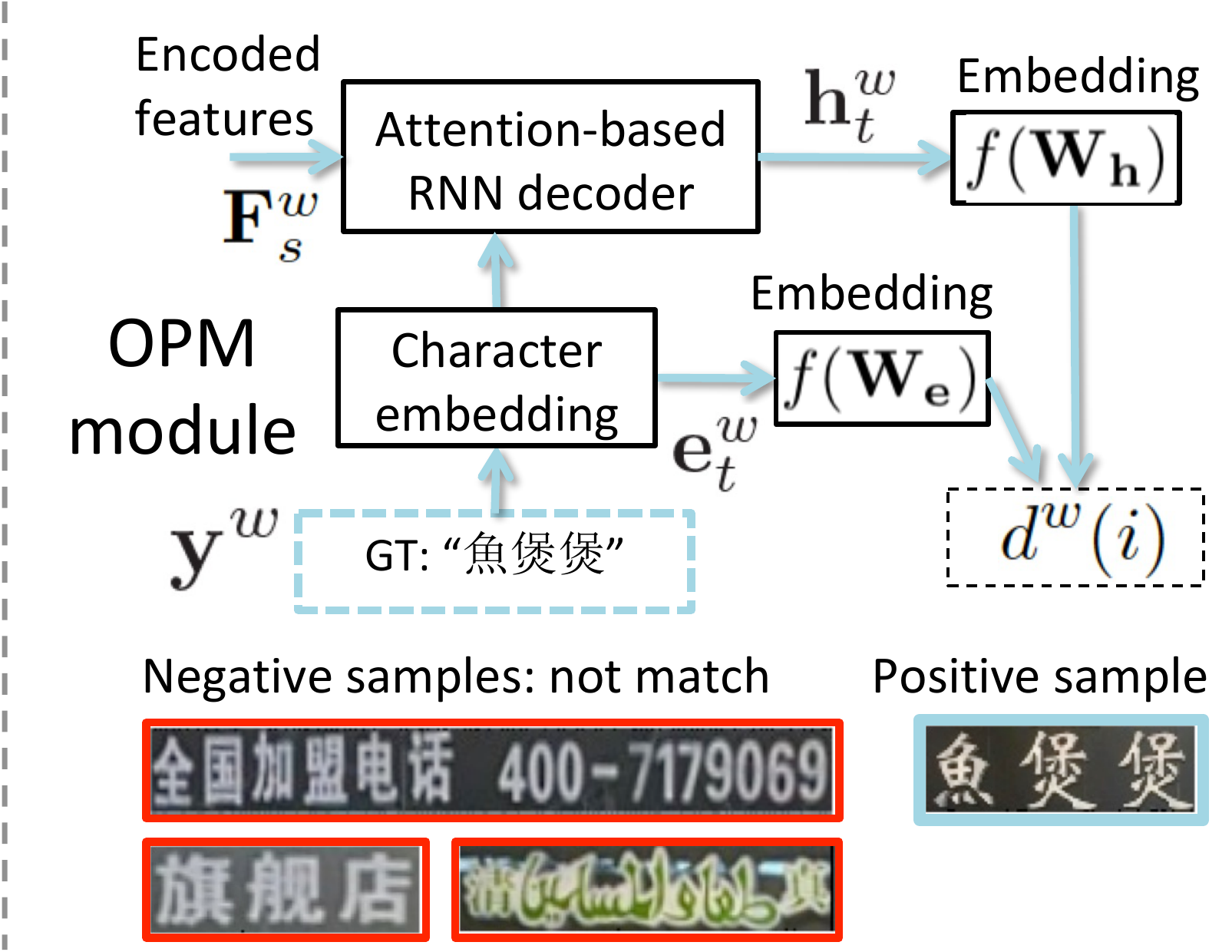}
   \end{center}\vspace{-1.5em}
\caption{(Left) the overall architecture of the proposed partially supervised end-to-end text reading model. (Right) online proposal matching.}\vspace{-1.2em}
\label{figure:overall_architecture}
\end{figure*}

\section{Partially Supervised Chinese Text Reading}
We first present the fully supervised end-to-end text reading model, and then introduce our partially supervised framework for reading Chinese text in the wild, consisting of the backbone network, text detection branch, perspective RoI~(Region-of-Interest) transform, text recognition branch and online proposal matching module, as shown in Fig. \ref{figure:overall_architecture}.
\vspace{-0.25em}

\subsection{End-to-End Chinese Text Reading}
\paragraph{Text Detection Branch:} Following the design of the existing text detector~\cite{zhou2017east}, we utilize the ResNet-50~\cite{he2016deep} as the standard base network and adopt FPN~(feature pyramid network) \cite{Lin_2017_CVPR} to fuse features at various scales, forming a shared backbone network for simultaneous text detection and recognition.
The text detection branch also consists of fully convolutional blocks, which jointly performs text/non-text classification and text location regression. Given the feature map $\mathbf{F}$ produced by the backbone network, text/non-text classification is conducted at each spatial location of $\mathbf{F}$ to compute its probability that belongs to text regions. To localize text instances, we directly adopt quadrangles to represent perspective text regions, 
%
predicting the offsets $\{(\Delta x_m, \Delta y_m) | m = 1,2,3,4\}$ between the location of each point and the four vertices of the text region that contains this point. 
During the training stage, the detection loss $L^{det}$ is defined as $L_{det} =  L_{loc} + \lambda L_{cls}$, where $L_{cls}$ is the dice loss for text/non-text classification, $L_{loc}$ is calculated as the smooth $L_{1}$ loss for location regression and $\lambda$ is a hyper-parameter that balances the two losses. In the testing stage, the detection branch applies thresholding to the predicted probabilities for text classification and performs NMS~(non-maximum suppression) on the selected spatial locations to generate quadrangle text proposals. 

\vspace{-1em}
\paragraph{Perspective RoI Transform:} Given quadrangle text proposals predicted by the detection branch, we employ the perspective RoI transform~\cite{sun2018textnet} to align the corresponding regions from the feature map $\mathbf{F}$ into small feature maps $\mathbf{F}_{p}$ rather than wrapping proposals in rotated rectangles~\cite{buvsta2017deep}\cite{liu2018fots}\cite{he2018end}. Each feature map $\mathbf{F}_{p}$ is kept in a fixed height with an unchanged aspect ratio. Different from English text recognition~\cite{li2017towards}\cite{liu2018fots}\cite{he2018end}\cite{sun2018textnet}, C-SVT text includes both horizontal and vertical lines. Therefore, we rotate each vertical line by $90$ degree in clockwise order if the aspect ratio (height/width) is above $1$, recognizing horizontal and vertical text lines in one recognition branch.
\vspace{-1em}
\paragraph{Text Recognition Branch:} Given the aligned small features $\mathbf{F}_{p}$, the text recognition branch is responsible for recognizing individual characters from these proposals.
Following the recent text recognition methods~\cite{shi2016robust}\cite{lee2016recursive}, we employ an attention based encoder-decoder framework in the recognition branch.
Taking the small feature map $\mathbf{F}_{p}$ as the input, the recognition branch extracts sequential text features $\mathbf{F}_s$ with stacked convolutional layers and RNN as the encoder. 
To recognize individual characters, we utilize the attention-based mechanism \cite{bahdanau2014neural} to translate the sequence of text features into the sequence of characters $\mathbf{y}$$=$$\{y_t\}_{t=1}^T$, where $y_t$ and $T$ are the character label and the length of word $\mathbf{y}$, respectively. In the decoding process at time step $t$, we utilize the attention mechanism with a RNN decoder to update the hidden state $\mathbf{h}_t$ and predict character $y_t$ given $y_{t-1}$ as the input. 
In our implementation, we use GRU~(Gated Recurrent Unit) as the recurrent unit for its ease-of-use. The probability of the output character label $y_t$ is then calculated with a fully connected layer and a softmax loss function. The training loss $L_{recog}$ of recognition branch is calculated over all text proposals with ground truth locations and labels, formulated as a fully supervised learning problem.

\subsection{Partially Supervised Learning}
Previous end-to-end trainable text reading models \cite{wang2012end,he2018end,li2017towards,liu2018fots,lyu2018mask} only utilize images in full annotations provided by the previous benchmarks~\cite{shi2017icdar2017,icpr-2018-mtwi,yuan2018chinese}. The improvement in performance of these models requires more fully annotated training data, which is extremely expensive and inefficient in annotations. To further scale up the amount of data while keeping it cost-effective, we aim to make full use of weakly annotated data, which is much cheaper. Therefore, we propose a text reading model with partially supervised learning, which enables to train full and weak annotations in a unified model to keep it simple yet effective. Existing weakly supervised learning methods for image classification \cite{mahajan2018exploring}, segmentation~\cite{Hu_2018_CVPR, shen2018bootstrapping} and text detection tasks~\cite{tian2017wetext} mainly focus on using image-level or bounding box as weak annotations, while 
end-to-end text reading aims to translate an image into a set of sequences, which is more challenging to solve. To address this problem, given only weak annotations in text labels without locations, we develop an Online Proposal Matching module (OPM) to select the best matched text proposals from weakly labeled images. This module is capable of spotting given keywords in an online fashion, and can be easily incorporated into the whole text reading model by adding recognition losses for end-to-end training. The overall architecture of the proposed partially learning model is illustrated in Fig. \ref{figure:overall_architecture}. 
\vspace{-0.5em}
\subsubsection{Online Proposal Matching}\label{subsubsection:opm}
To spot keywords from weakly annotated images in a unified framework, the OPM module can be integrated into the proposal-based supervised learning framework by sharing parts of the learned parameters. Given a weakly labeled image $\mathbf{I}^w$, the OPM module aims to locate the text regions corresponding to the keyword annotation $\mathbf{y}^w$. We first utilize the detection branch of the fully supervised model to generate a set of text proposals $\{P^{w}(i)\}_{i = 1}^N$, where $N$ is the number of the predicted text proposals. The feature map of each proposal is then extracted by perspective RoI transform and encoded as sequential features $\mathbf{F}_s^{w}$ by the CNN-RNN encoder in the text recognition branch. Furthermore, to calculate the similarity between features $\mathbf{F}_s^{w}$ and the weakly labeled keyword $\mathbf{y}^w$, we utilize an attention-based RNN decoder in the OPM module to compute decoder states $\{\mathbf{h}^{w}_t\}_{t = 1}^{T^w}$ given $\mathbf{y}^w$ as the input as shown in Fig. \ref{figure:overall_architecture}. Note that the attention-based RNN decoder shares the same parameters and character embedding layer with that of the recognition branch, and $T^{w}$ is the number of time steps, which is also the length of keyword $\mathbf{y}^w$. To select correct proposals that contain the keyword, the OPM module directly computes the Euclidean distance $d^{w}(i)$ in the embedding space $f(\cdot)$ between the decoder states $\{\mathbf{h}^{w}_t\}_{t = 1}^{T^w}$ for each text proposal and the character embeddings $\{\mathbf{e}_{t}^{w}\}_{t=1}^{T^w}$ of keyword $\mathbf{y}^w$ as\vspace{-0.5em}
\begin{equation}
   d^w(i) = \frac{1}{T^w}\sum_{t=1}^{T^w}||f(\mathbf{h}^w_t, \mathbf{W}_h) - f(\mathbf{e}^w_t, \mathbf{W}_e)||,
\end{equation}
where $\mathbf{W}_h$ and $\mathbf{W}_e$ are parameters to encode $\mathbf{h}^w_t$ and $\mathbf{e}^w_t$ in the embedding space, respectively. During the training process, the OPM module is trained using a pairwise loss\vspace{-0.75em}
\begin{equation}\label{equation:l_opm}
L_{opm} = \frac{1}{N}\sum_{i = 1}^{N}[s^w(i)]^2,
\end{equation}
where $s^w(i) = d^w(i)$ if the text proposal $P^w(i)$ is a positive sample that matches the keyword $\mathbf{y}^w$, otherwise $s^w(i) = max(0, 1-d^w(i))$. To train OPM, we generate positive and negative samples by checking the IoU between $P^w(i)$ and the selected ground-truth keyword region  (see Sec. 4.3).

\vspace{-0.5em}
\subsubsection{Fully and Weakly Supervised Joint Training}\label{subsubsection:psl}
As illustrated in Fig.~\ref{figure:overall_architecture}, the partially supervised learning framework for reading Chinese text in natural images is composed of two parts, namely fully and weakly supervised text reading. For the fully supervised text reading, the training images come from the fully annotated dataset of C-SVT and the training loss $L_{full}$ is computed as\vspace{-0.5em}
\begin{equation}\vspace{-0.5em}
L _{full}= L_{det} + \beta L_{recog},
\end{equation}
where $\beta$ is the parameter to trade-off these two losses. For the weakly supervised Chinese text reading, we use the weakly annotated images together with the proposed OPM module. The training loss $L_{recog}^{w}$ is formulated as\vspace{-0.5em}
\begin{equation}\label{eqn:ws}
L_{recog}^{w}=\frac{1}{\sum_{i=1}^{N}m(i)}\sum_{i=1}^{N}m(i)l_{recog}^w(i),
\end{equation}
where $m(i)=1$ if $d^w(i) \le \tau$,~otherwise $m(i)=0$ and a threshold $\tau$ is used to select the matched text proposals. The recognition loss $l^{w}_{recog}(i)$ of the $i$-th text proposal is defined as the negative log-likelihood function
\begin{equation}\label{eqn:wrecog}
l_{recog}^w(i) = - \frac{1}{T^w}\sum_{t=1}^{T^w} \log p(\mathbf{y}^{w}_{t} | \mathbf{y}^w_{t-1}, \mathbf{h}^{w}_{t-1}, \mathbf{c}^{w}_t),
\end{equation}
where $\mathbf{c}_t^{w}$ denotes the context vector at time $t$ calculated by attention mechanism.
The total loss for the partially supervised learning framework is therefore computed as\vspace{-0.5em}
\begin{equation}\label{eqn:ps}
L_{total} = L_{det} + \beta (L_{recog} + L_{recog}^{w})
\end{equation}
for fully and weakly supervised joint training.

\subsection{Training Pipeline}
The fully supervised Chinese text reading model is pretrained using the VGG synthetic dataset \cite{gupta2016synthetic} and then finetuned on the fully annotated data of C-SVT. The training process of the proposed partially supervised framework is built upon the fully supervised text reading model and can be divided into two stages:

\textbf{Stage One}: We first train the OPM module as described in Sec. \ref{subsubsection:opm} by fixing the well trained fully supervised part. As we do not have the ground truth locations of the keyword regions in weakly labeled images, we create pseudo weakly labeled training samples generated from the fully annotated images. Given a fully annotated image, we randomly select one of the labeled text instances as the keyword region and generate a set of text proposals. To train the OPM module (see Eqn. (\ref{equation:l_opm})), we compute the IoU between each generated proposal and the selected keyword region and choose those proposals whose IoUs are smaller than $0.5$ as the negative examples. We directly use the ground truth locations of the selected keyword regions as positive examples .

\textbf{Stage Two}: With the trained OPM module, we further train the whole model with the partially supervised loss $L_{total}$ (see Eqn. (\ref{eqn:ps})). In this stage, we feed both fully and weakly annotated samples into the partially supervised model, which is end-to-end trainable.
 

\begin{table*}[!ht]
\renewcommand{\arraystretch}{1.1}
\tabcolsep=0.20em
\centering
\scriptsize 
\caption{The performance of the end-to-end Chinese text reading models on C-SVT . `PSL' denotes the proposed partially supervised learning algorithm.}\label{table:chi-svt}\vspace{-1em}
\begin{tabular}{|l|c||c|c|c||c|c|c|c||c|c|c||c|c|c|c|}
  \hline
  \multirow{3}{*}{Method} &
  \multirow{3}{*}{Training data} &
  \multicolumn{7}{c||}{Valid} &
  \multicolumn{7}{c|}{Test} \\
  \cline{3-16}
  & & \multicolumn{3}{c||}{Detection} &
  \multicolumn{4}{c||}{End-to-end} &
  \multicolumn{3}{c||}{Detection} &
  \multicolumn{4}{c|}{End-to-end} \\ 
  \cline{3-16} 
                                &                            & R \%  & P \%  & F \%  & R \%  & P \%  & F \%  & AED   & R \%  & P \%  & F \%  & R \%  & P \%  & F \%  & AED   \\
  \hline
  \multirow{1}{*}{EAST\cite{zhou2017east}+Attention\cite{shi2016robust}}   & Train            &  71.74 & 77.58 & 74.54 & 23.89  & 25.83 & 24.82 & 22.29  & 73.37& 79.31 & 76.22 & 25.02 & 27.05 & 25.99 & 21.26   \\
  \hline
  \multirow{1}{*}{EAST\cite{zhou2017east}+CRNN\cite{shi2017end}}      & Train               &  71.74 & 77.58 & 74.54 & 25.78 & 27.88 & 26.79 & 20.30  & 73.37& 79.31 & 76.22  & 26.96 & 29.14 & 28.0 & 19.25  \\
  \hline\hline
  \multirow{3}{*}{End2End}      & Train                      & 72.70 & 78.21 & 75.35 & 26.83 & 28.86 & 27.81 & 20.01  & 74.60 & 80.42 & 77.40 & 27.55 & 29.69 & 28.58 & 19.68  \\
  \cline{2-16}
       & Train + $4.4$K Extra Full                & 72.98 & 78.46 & 75.62 & 28.03 & 30.13 & 29.04 & 19.62  & 74.95 & 80.84 & 77.79 & 28.77 & 31.03 & 29.85 & 19.06 \\
  \cline{2-16}
        & Train + $10$K Extra Full                & 73.23 & 76.69 & 74.92 & 29.91 & 31.32 & 30.60 & 18.87  &  75.13 & 78.82 & 76.93 & 30.57 & 32.07 & 31.30 & 18.46\\
  \cline{1-16}
  \multirow{4}{*}{End2End-PSL}  & Train + $25$K Weak        & 72.93 & 79.37 & 76.01 & 29.44 & 32.04 & 30.68 & 19.47 & 74.72 & 81.39 & 77.91 & 30.18 & 32.87 & 31.46 & 18.82 \\
  \cline{2-16}
                                & Train + $50$K Weak	         & 73.09 & 79.36 & 76.10 & 29.96 & 32.53 & 31.19 & 19.20  & 74.80 & 81.32 & 77.93 & 30.56 & 33.22 & 31.83 & 18.72  \\
  \cline{2-16}
                                & Train + $100$K Weak	     & 73.17 & 78.50 & 75.74 & 30.55 & 32.78 & 31.63 & 18.97 & 75.04 & 80.41 & 77.63 & 31.19 & 33.43 & 32.27 & 18.28  \\
  \cline{2-16}
                                & Train + $200$K Weak	     & 73.26 & 78.64 & 75.85 & 31.31 & 33.61 & 32.41 & 18.54 & 75.14 & 80.68 & 77.81 & 32.01 & 34.38 & 33.15 & 18.12 \\ 
  \cline{2-16}
                                & Train + $400$K Weak	     & \bf{73.31} & \bf{79.73} & \bf{76.38} & \bf{31.80} & \bf{34.58} & \bf{33.13} & \bf{18.14} & \bf{75.21} & \bf{81.71} & \bf{78.32} & \bf{32.53} & \bf{35.34} & \bf{33.88} & \bf{17.59}  \\

  \hline
\end{tabular}\vspace{-0.6em}
\end{table*}

\begin{figure*}[t]
\begin{minipage}{\linewidth}
   \begin{center}
     \includegraphics[width=0.95\linewidth]{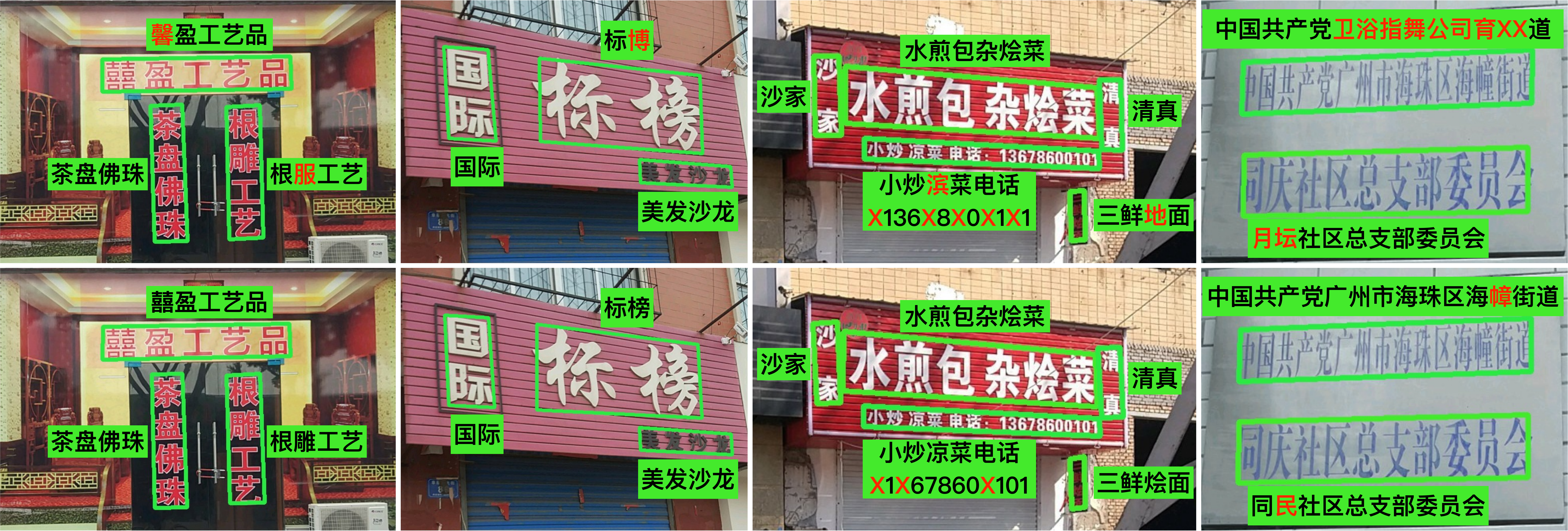}
   \end{center}
\end{minipage}\vspace{-0.8em}
\caption{Qualitative comparisons between the fully and partially supervised text reading models. Visual results on the top and bottom rows are predicted by the fully and partially supervised models, respectively. The incorrectly recognized characters are marked in red. `X' denotes the prediction of the corresponding character is missing. }\label{figure:psm_examples}\vspace{-1.2em}
\end{figure*}

\section{Experiments}
In this section, we conduct experiments on our C-SVT dataset to demonstrate the effectiveness of the proposed partially supervised learning approach. Note that in all the experiments, we denote Recall, Precision and F-score as `R', `P' and `F' respectively. The average edit distance
 defined in ICDAR 2017-RCTW \cite{shi2017icdar2017} 
 is 
 denoted as `AED'
. Higher `R', `P', `F'
   mean better performance while lower `AED' means better performance.

\subsection{Implementation Details}
When using the fully annotated data of C-SVT, data augmentation is performed to improve the robustness of our text reading model. First, we randomly chose scale factors from $[0.5, 1, 2.0, 3.0]$ to rescale the original training images. Then, we randomly crop samples from the rescaled images and resize their longer sides to $512$ pixels. Finally, we pad the images to $512\times512$ with zero values. In the RoI transform layer, we set the height and the maximum width of the warped feature map to $8$ and $64$ respectively. If the width of the feature map is smaller than $64$, we pad it using zero values. Otherwise, we resize it with bilinear interpolation to set the width to $64$. All the weakly labeled images in Stages 1 and 2 of the partially supervised framework are resized to $512\times512$ with paddings. All of our experiments are conducted on eight NVIDIA TESLA P40 GPUs. For the pretraining and finetuning of our model, the batch size is $16$ per GPU and the number of text proposals in a batch is set to $32$ per GPU. For the partially supervised learning framework, we have two data streams: fully and weakly annotated training images. We set each batch size equal to $8$. Throughout the whole training process, we use Adam as the optimization algorithm and the learning rate is set to $10^{-4}$. The parameters $\lambda$ and $\beta$ are set to $0.01$ and $0.02$ as defaults, respectively.

\subsection{Quantitative and Qualitative Results}
Following the evaluation protocols, the results of the text detection and end-to-end recognition on C-SVT are shown in Tab.~\ref{table:chi-svt}. Note that `End2End' denotes the end-to-end trainable model with only full annotations, and `End2End-PSL'  means the proposed end-to-end partially supervised learning model trained on full and weak annotations. 
By taking advantages of the large-scale weakly annotated data, the partially supervised model can achieve considerably better performance in end-to-end text reading. Specifically, compared with `End2End' trained with full annotation of C-SVT, the performance of `End2End-PSL' with $400$K weakly annotated data increases by $5$\% in terms of recall, precision and F-score. Compared with other approaches, e.g., EAST \cite{zhou2017east}+CRNN \cite{shi2017end} and EAST \cite{zhou2017east}+Attention \cite{shi2016robust}, trained on C-SVT datasets, the proposed partially supervised algorithm surpasses these methods in terms of both F-score and 
AED
metrics. 
From the examples shown in Fig.~\ref{figure:psm_examples}, we notice that our partially supervised model shows better visual results than the fully supervised one. 

We conduct experiments to explore how the performance of our text reading model is affected by the amount of the weakly annotated images. As illustrated in Tab. \ref{table:chi-svt}, when adding more weakly annotated images, the end-to-end text reading performance of our model on the test set can be improved continuously from $28.58\%$ to $33.88\%$ in F-score. To further validate the effectiveness of weak annotations, we also randomly select $10$K images from weakly labeled images and label them in full annotations as `Extra Full' data to train the `End2End' models. Following the annotation cost counting in man-hours (see Sec. 3.2), the labeling cost of $4.4$K full annotations approximately equals to that of $400$K weak annotations, and `End2End-PSL' (Train + $400$K Weak) has shown considerable end-to-end performance improvement by 
$4.03$\% in F-score and $1.47$\% in AED 
over `End2End' (Train + Extra Full $4.4$K) on the test set. We also notice that the end-to-end performance of `End2End-PSL' (Train + $50$K Weak) in 
F-score and AED 
is compatible with that of `End2End' (Train + $10$K Extra Full), while the labeling cost of $50$K weak annotations is only $\frac{1}{12}$ of that of $10$K full annotations, which further demonstrates the effectiveness of the weakly annotated data. 

\begin{figure*}[t]
\begin{minipage}{\linewidth}
   \begin{center}
     \includegraphics[width=0.95\linewidth]{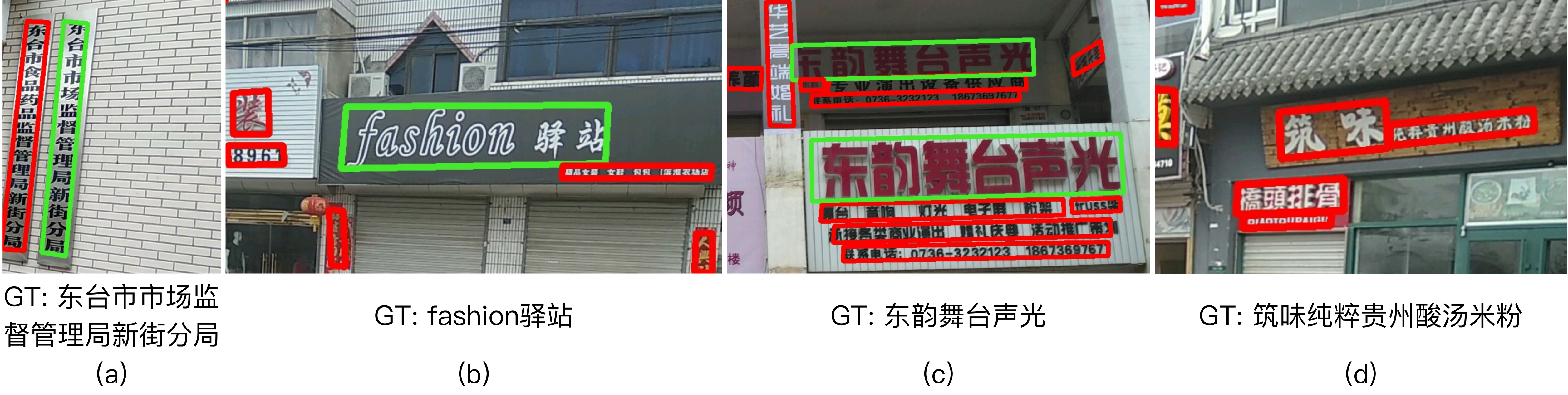}
   \end{center}
\end{minipage} \vspace{-1.2em}
\caption{Matching examples generated by OPM module. The green and red quadrangles represent the matched and mis-matched text proposals with respect to the ground-truth keyword, respectively. }\label{figure:opm_examples}\vspace{-1.2em}
\end{figure*}


\subsection{Comparisons with Other Approaches}

To compare with other approaches, we also conduct experiments on the ICDAR 2017-RCTW dataset~\cite{shi2017icdar2017}, which contains $8,034$ images for training and $4,229$ for testing labeled in full annotations. We train our baseline text reading model `End2End'  with VGG synthetic data pretraining and finetune on the train set of ICDAR 2017-RCTW.  To validate the effectiveness of the proposed partially supervised learning algorithm, we use the train set of ICDAR 2017-RCTW and $400$K weakly annotated data of C-SVT to train `End2End-PSL' to further improve the end-to-end performance. The detection and end-to-end recognition results are shown in Tab.~\ref{table:rctw}. Note that `MS' denotes multi-scale testing of the trained models. 
\begin{table}[t]
\renewcommand{\arraystretch}{1.1}
\tabcolsep=0.36em
\centering
\scriptsize 
\caption{Comparisons with other methods on ICDAR 2017-RCTW. Note that `MS' denotes testing in multi-scales and `PSL' denotes our partially supervised learning algorithm. 
}\label{table:rctw}\vspace{-1em}
\begin{tabular}{|l||c|c|c||c|c|c|c|}
  \hline
   \multirow{2}{*}{Method} & \multicolumn{3}{c||}{Detection} & \multicolumn{2}{c|}{End-to-end} \\
   \cline{2-6}
                                             & R \% & P \% & F \% & AED                 & Norm\% \\
  \hline
  RCTW-baseline \cite{shi2017icdar2017}      & 40.4 & 76   & 52.8 & $25.62^{+}$              & *      \\
  \hline
  EAST(ResNet-50)~\cite{zhou2017east}        & 47.8 & 59.7 & 53.1 & *                   & *      \\
  \hline
  RRD \cite{liao2018rotation}                & 45.3 & 72.4 & 55.7 & *                   & *      \\
  \hline
  RRD-MS \cite{liao2018rotation}             & \bf{59.1} & 77.5 & 67.0 & *                   & *      \\
  \hline
  Border(ResNet)-MS \cite{xue2018accurate}   & 58.5 & 74.2 & 65.4 & *                   & *      \\
   \hline
  Border(DenseNet)-MS \cite{xue2018accurate} & 58.8 & 78.2 & 67.1 & *                   & *      \\
  \hline\hline
  End2End                                    & 47.2 & \bf{82.8} &  60.1 & 27.5 & 72.9 \\
  \hline
  End2End-PSL                                & 47.4 & 82.5 & 60.2  & 24.2 & 76.2 \\
  \hline
  End2End-MS                                 & 57.2 & 82.4 & 67.5  & 26.2      & 73.5       \\
  \hline
  End2End-PSL-MS                             & 57.8 & 81.7 & \bf{67.7}  & \bf{22.1} & \bf{77.7}\\
  \hline
\end{tabular}
\vspace{-1.4em}
\end{table}
Compared with the previous methods, e.g., RRD \cite{liao2018rotation} and Border \cite{xue2018accurate} , our baseline text reading models (`End2End' and `End2End-MS') show slightly better performance in F-score for detection, which is tested in single and multiple scales, respectively. Note that the end-to-end baseline of ICDAR 2017-RCTW \cite{shi2017icdar2017} marked with ‘+’ used a large synthetic dataset with a Chinese lexicon to pretrain the recognition model. 
From Tab.~\ref{table:rctw}, it can be seen that by leveraging weak annotations of C-SVT, the proposed `End2End-PSL' can surpass the fully supervised model `End2End' by mainly boosting the recognition performance, reducing AED by $3.3$ per image. In multi-scale testing, we can further reduce the average distance to $22.1$ to provide a new state-of-the-art result, which demonstrates the effectiveness and generalization of the proposed partially supervised learning algorithm.

\begin{table}[t]
\renewcommand{\arraystretch}{1.1}
\centering
\scriptsize
\caption{The performance of OPM in terms of recall~(R) and precision~(P).}\label{table:opm}\vspace{-1.2em}
\begin{tabular}{|c||c|c|c||c|c|c|c|}
  \hline
  $\tau$ & IoU & R \%& P \% & IoU & R \%& P \% \\
  \hline
  0.05   &  \multirow{3}{*}{0.5} & 45 & 98 &  \multirow{3}{*}{0.7} & 45 & 96  \\
  \cline{1-1}\cline{3-4}\cline{6-7}
  0.1    &                       & 57 & 97 &                       & 56 & 94  \\
   \cline{1-1}\cline{3-4}\cline{6-7}
  0.2    &                       & 66 & 92 &                       & 64 & 89  \\
  \hline
\end{tabular}\vspace{-1.4em}
\end{table}

\subsection{Effectiveness of Model Design}

\paragraph{Online Proposal Matching:} To validate the effectiveness of OPM, we randomly select $500$ images from the weakly annotated data of C-SVT and manually evaluate the accuracy in  spotting keywords. Tab. \ref{table:opm} shows the performance of the OPM module by using different thresholds $\tau$ to generate matched text proposals. We observe that the OPM module can have an acceptable recall and a high precision when the threshold $\tau$ equals to $0.1$, so we choose $\tau=0.1$ in all the experiments to train the partially supervised model.

As the matching examples shown in Fig.~\ref{figure:opm_examples}, the proposed OPM module is able to precisely select the correct text proposals according to the corresponding keywords in the weakly labeled images. Even when the keyword appears several times in an image, OPM can still localize all the corresponding text regions (see Fig. \ref{figure:opm_examples}(c)). The recall and precision of our OPM are $57\%$ and $97\%$, respectively. The main reason for the low recall is that the OPM module fails when the detection branch splits some keywords into multiple proposals, as shown in Fig.~\ref{figure:opm_examples}(d). 

\begin{table}[t]
\renewcommand{\arraystretch}{1.1}
\centering
\scriptsize 
\caption{The performance of the recognition branch of our Chinese text reading models.}\label{table:recog}\vspace{-1.4em}
\begin{tabular}{|c|c|c|c|c|c|c|c|}
  \hline
  Method                       & Training Data       & Accuracy \% & AED      \\
  \hline\hline
  End2end                      & Train               & 40.20       & 13.52   \\
  \hline
  \multirow{1}{*}{End2end-PSL} 
                               & Train + $400$K Weak  & \bf{48.33 }  & \bf{10.79}   \\
  \hline
\end{tabular}\vspace{-1.4em}
\end{table}

\vspace{-.5em}
\paragraph{Effectiveness of weak Annotations on Recognition:} 
We also evaluate the performance of the recognition branch of `End2End-PSL' on the C-SVT test set. In this experiment, we directly use the ground truth location of each text region as the input of perspective RoI transform. As shown in Tab.~\ref{table:recog}, the accuracy of the recognition branch can be improved by a large margin when adding $400$K weakly labeled data for training. Therefore, the weak annotated data of C-SVT plays an important role in boosting the recognition ability. 

\vspace{-0.5em}
\section{Conclusions}
We have developed a new large-scale Chinese text reading benchmark, i.e., Chinese Street View Text, which is the largest one compared with existing Chinese text reading datasets. To scale up training samples while keeping the labeling process cost-effective, we annotated images in full and $400$K weak labels. We proposed an online proposal matching module to train the end-to-end text reading model with partially supervised learning, simultaneously learning from fully and
weakly annotated data in a unified framework. Experiments on C-SVT have shown its superior performance and the proposed model with large-scale weak annotations can improve the end-to-end results by $4.03$\% in F-score over the fully supervised model at the same labeling cost, and achieved the state-of-the-art results on the ICDAR 2017-RCTW dataset.

\clearpage
\noindent\textbf{\large Supplemental Materials}\\

\noindent\textbf{1. Experiments on English dataset}
We train `End2End-PSL' with ICDAR-13, ICDAR-15 and MLT-17 Latin datasets as full annotations, and generate the weakly labeled data from COCO-Text dataset. In single-scale testing for fair comparisons, `End2End-PSL' outperforms `End2End'  by $3.24\%$ / $3.66\%$ under generic conditions on the ICDAR-15 test set, surpassing or achieving compatible results with existing methods.

\begin{table}[ht]
\begin{center}
\footnotesize 
\caption{End-to-end results on ICDAR 2015.}
\vspace{-0.5em}
\label{tab:icdar-15}
\renewcommand{\arraystretch}{1.}
\tabcolsep=0.3em
\begin{tabular}{|c|c|c|c|c||c|c|c|}
  \hline
  \multirow{2}{*}{Single-scale testing} & Det. &
  \multicolumn{3}{c||}{End-to-end} &
  \multicolumn{3}{c|}{Word spotting} \\ 
  \cline{2-8} 
  &  F & S & W & G & S & W & G \\
  \hline
  HUST MCLAB [33][34] & 75 & 67.86 & * & * & 70.57 & * & * \\
  \hline
  Deep TextSpotter [ICCV'17] & * & 54.0 & 51.0 & 47.0 & 58.0 & 53.0 & 51.0 \\
  \hline 
  Mask TextSpotter [ECCV'18]  & 86 & 79.3 &  73.0  &  \bf{62.4}  &79.3 &  74.5 & \bf{64.2}\\
  \hline
  FOTS [CVPR'18]         & \bf{87.99} & 81.09 &  75.9  &  60.8  & 84.64 &  79.32  & 63.29\\
  \hline\hline
  End2End                     & 87.01 & 81.08 &  75.09 &  59.93  & 84.62 &  78.35  & 60.92\\
  \hline
  End2End-PSL & 87.24 & \bf{81.18} &  \bf{76.25}  &  62.29  & \bf{84.77} &  \bf{79.74}  & 63.91\\
   \hline
\end{tabular}
\end{center}
\end{table}

\noindent\textbf{2. Comparisons against the CTW dataset}

We train the fully supervised end-to-end model `End2End' with CTW and CSVT Full (the fully annotated part), and evaluate results on the ICDAR 2017-RCTW test set. 
End2End trained with CSVT fully annotated data surpasses the model trained with CTW in AED/Norm by $6.8$ and $7.4\%$.

\begin{table}[ht]
\renewcommand{\arraystretch}{1.0}
\tabcolsep=0.36em
\centering\vspace{-0.5em}
\small
\caption{Comparisons between fully annotated datasets.}
\label{table:ctw_rctw}\vspace{-0.5em}
\begin{tabular}{|c|c||c|c|c||c|c|c|}
  \hline
   \multirow{2}{*}{Testing} & \multirow{2}{*}{Training data} & \multicolumn{3}{c||}{Detection} & \multicolumn{2}{c|}{End-to-end} \\
   \cline{3-7}
                   &    & R \% & P \% & F \% & AED                 & Norm\% \\
   \cline{1-7}
    ICDAR 2017 &  CTW [43]  & 30.1 & 78.1   & 43.41 & 33.6              & 66.4    \\
   \cline{2-7}
   RCTW & CSVT Full   & 45.29 & 81.43 &  58.21 &  26.8             & 73.8    \\
    \hline
  \end{tabular}
\end{table}

\clearpage
{\small
\bibliographystyle{ieee_fullname}
\bibliography{egbib,dataset_egbib}
}

\end{document}